# Modeling and Control of CSTR using Model based Neural Network Predictive Control

Piyush Shrivastava
Assistant Professor,
Electrical & Electronics Engineering Department,
Takshshila Institute of Engineering & Technology,
Jabalpur, Madhya Pradesh, India
e-mail: piyush247@gmail.com

*Abstract*— this paper presents a predictive control strategy based on neural network model of the plant is applied to Continuous Stirred Tank Reactor (CSTR). This system is a highly nonlinear process; therefore, a nonlinear predictive method, e.g., neural network predictive control, can be a better match to govern the system dynamics. In the paper, the NN model and the way in which it can be used to predict the behavior of the CSTR process over a certain prediction horizon are described, and some comments about the optimization procedure are made. Predictive control algorithm is applied to control the concentration in a continuous stirred tank reactor (CSTR), whose parameters are optimally determined by solving quadratic performance index using the optimization algorithm. An efficient control of the product concentration in cstr can be achieved only through accurate model. Here an attempt is made to alleviate the modeling difficulties using Artificial Intelligent technique such as Neural Network. Simulation results demonstrate the feasibility and effectiveness of the NNMPC technique.

## I. INTRODUCTION

One of the main aims in industry is to reduce operating costs. This implies improvements in the final product quality, as well as making better use of the energy resources. Advanced control systems are in fact designed to cope with these requirements. Model based predictive control (MBPC) [1,2] is now widely used in industry and a large number of implementation algorithms due to its ability to handle difficult control problems which involve multivariable process interactions, constraints in the system variables, time delays, etc. The most important advantage of the MPC technology comes from the process model itself, which allows the controller to deal with an exact replica of the real process dynamics, implying a much better control quality. The inclusion of the constraints is the feature that most clearly distinguishes MPC from other process control techniques, leading to a tighter control and a more reliable controller.

Another important characteristic, which contributes to the success of the MPC technique, is that the MPC algorithms consider plant behavior over a future horizon in time. Thus, the effects of both feedforward and feedback disturbances can be anticipated and eliminated, fact which permits the controller to drive the process output more closely to the reference trajectory. The classical MBPC algorithms use linear models of the process to predict the output of the process over a certain horizon, and to evaluate a future sequence of control signals in order to minimize a certain cost function that takes account of the future output prediction errors over a reference trajectory, as well as control efforts. Although industrial processes especially continuous and batch processes in chemical and petrochemical plants usually contain complex nonlinearities, most of the MPC algorithms are based on a linear model of the process and such predictive control algorithms may not give rise to satisfactory control performance [3, 4]. Linear models such as step response and impulse response models are preferred, because they can be identified in a straightforward manner from process test data. In addition, the goal for most of the applications is to maintain the system at a desired steady state, rather than moving rapidly between different operating points, so a precisely identified linear model is sufficiently accurate in the neighborhood of a single operating point. As linear models are reliable from this point of view, they will provide most of the benefits with MPC technology. Even so, if the process is highly nonlinear and subject to large frequent disturbances; a nonlinear model will be necessary to describe the behavior of the process. Also in servo control problems where the operating point is frequently changing, a nonlinear model of the plant is indispensable. In situations like the ones mentioned above, the task of obtaining a high-fidelity model is more difficult to build for nonlinear processes.

In recent years, the use of neural networks for nonlinear system identification has proved to be extremely successful [5-9]. The aim of this paper is to develop a nonlinear control technique to provide high-quality control in the presence of nonlinearities, as well as a better understanding of the design process when using these emerging technologies, i.e., neural network control algorithm. The combination of neural networks and model-based predictive control seems to be a good choice to achieve good performance in the control. In this paper, we will use an optimization algorithm to minimize the cost function and obtain the control input. The paper analyses a neural network based nonlinear predictive controller for a Continuous Stirred Tank Reactor (CSTR), which is a highly nonlinear process. The procedure is based on construction of a neural model for the process and the proper use of that in the optimization process.

This paper begins with an introduction about the predictive control and then the description of the nonlinear predictive control and the way in which it is implemented. The neural model and the way in which it can be used to predict the behavior of the CSTR process over a certain prediction horizon are described, and some comments about the optimization procedure are made. Afterwards, the control aims, the steps in the design of the control system, and some simulation results are discussed.

## II. PREDICTIVE CONTROL

The predictive controller, in summary, is characterized by computing future control actions based on output values predicted by a model, with vast literature and academic and industrial interest (Clarke, 1987; Garcia et all, 1989; Arnaldo, 1998) [4]. This section presents the concepts of predictive control based on NPC, using the usual optimization functions and control laws, applied to the conventional predictive controllers.

### A. Optimization functions

The optimization function, usually represented by the index J, represents the function that the control action tries to minimize. In an intuitive way, the error between the plant output and the desired value is the simplest example of an optimization function, and it is expressed by:

$$J = y_{ref}(k) - y(k) = e(k) \quad (1)$$

Where:
$y(k)$ represent the plant output
$y_{ref}(k)$ represent the desired response
$e(k)$ represent the estimation error
$k$ is the sample time

One of the most usual optimization functions is based on the square error and it is represented as:

$$J = \left[y_{ref}(k) - y(k)\right]^2 = \left[e(k)\right]^2 \quad (2)$$

But the optimization index can take forms of more complex functions. For predictive controllers, whose models are capable to predict $N$ steps ahead, the simple application of the square error approach can present satisfactory results. This case admits that the optimization function is not limited to an only point, but an entire vector of $N$ predicted errors. It seeks to optimize the whole trajectory of the future control actions in a horizon of $N$ steps ahead.

$$J = \sum_{j=1}^{N}\left[y_{ref}(k+j) - \hat{y}(k+j)\right]^2 = \sum_{j=1}^{N}\left[e(k+j)\right]^2 \quad (3)$$

More complex optimization functions can consider the control effort. It is the specific case of GPC (Generalized Predictive Control), where the optimization index J can be expressed as:

$$J = \sum_{j=N_1}^{N_Y}\left[y_{ref}(k+j) - \hat{y}(k+j)\right]^2 + \sum_{j=1}^{N_U}\alpha(j).\left[\Delta u(k+j)\right]^2 \quad (4)$$

where:
$y(k)$ - is the output plant estimation at instant $= k$
$\Delta u$ - is the control action increment.
$N_1$ - is the minimum horizon of prediction.
$N_U$ - is the control horizon.
$N_Y$ - is the maximum horizon of prediction.

The objective of the control problem is to minimize the index J, with respect to the control actions, looking for the points where the first order differential is null.

## III. NEURAL NETWORK PREDICTIVE CONTROL

By the knowledge of the identified neural model of the nonlinear plant which is capable of doing multi step ahead predictions, Predictive control algorithm is applied to control nonlinear process. The idea of predictive control is to minimize cost function, $J$ at each sampling point:

$$J(t,U(k)) = \sum_{t=N_1}^{N_2}\left[r(k+i) - \hat{y}(k+i)\right]^2 + \sum_{i=1}^{N_u}\rho\left[\Delta u(k+i-1)\right]^2 \quad (5)$$

With respect to the $N_u$ future controls,

$$U(k) = [u(k)....u(k+N_u-1)]^T \quad (6)$$

and subject to constraints:

$$N_u \leq i \leq (N_2 - n_k) \quad (7)$$

Using the predictive control strategy with identified NARX model (NNMPC) it is possible to calculate the optimal control sequence for nonlinear plant. Here, term $r(k+i)$ is the required reference plant output, $\hat{y}(k+i)$ is predicted NN model output, $\Delta u(k+i-1)$ is the control increment, $N_1$ and $N_2$ are the minimum and maximum prediction (or cost) horizons, $Nu$ is the control horizon, and $\rho$ is the control penalty factor[4].

The predictive control approach is also termed as a receding horizon strategy, as it solves the above-defined optimization problem [5] for a finite future, at a current time and implements the first optimal control input as the current control input. The vector $u = [\Delta u(k), \Delta u(k+1),...\Delta u(k + Nu-1)]$ is calculated by minimizing cost function, $J$ at each sample $k$ for selected values of the control parameters $\{N_1, N_2, Nu, \rho\}$.

These control parameters defines the predictive control performance. $N_1$ is usually set to a value 1 that is equal to the time delay, and $N_2$ is set to define the prediction horizon i.e. the number of time-steps in the future for which the plant response is recursively predicted.

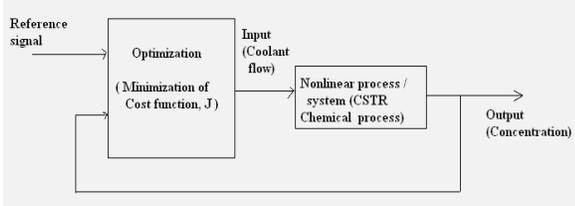

Figure 1: NNMPC principle applied to CSTR chemical process

The minimization of criterion, *J* in NNMPC is an optimization problem minimized iteratively. Similar to NN training strategies, iterative search methods are applied to determine the minimum.

$$\theta^{(i+1)} = \theta^{(i)} + \mu^{(i)} \cdot d^{(i)} \quad (8)$$

where, $\theta^{(i)}$ specifies the current iterate (number 'i'), $d^{(i)}$ is the search direction and $\mu^{(i)}$ is the step size. Various types of algorithms exist, characterized by the way in which search direction and step size are selected. In the present work Newton based Levenberg–Marquardt (LM) algorithm is implemented. The search direction applied in LM algorithm is:

$$(\hat{H}[U^i(t)] + \lambda^i I) d^i = -G[U^i(t)] \quad (9)$$

with Gradient vector and Hessian matrix as:

$$\begin{aligned} G[U^i(t)] &= \frac{\partial J(t, U(t))}{\partial U(t)} \Big|_{U(t)=U^i(t)} \\ &= -2\varphi^T[U^i(t)]\tilde{E}(t) + 2\rho \frac{\partial \tilde{U}(t)}{\partial U(t)} \tilde{U}(t) \Big|_{U(t)=U^i(t)} \end{aligned} \quad (10)$$

$$\begin{aligned} H[U^i(t)] &= \frac{\partial^2 J(t, \tilde{U}(t))}{\partial U(t)^2} \Big|_{U(t)=U^i(t)} \\ &= \frac{\partial}{\partial U(t)} \left( \frac{\partial \hat{Y}(t)}{\partial U(t)} E(t) \right) + 2\rho \frac{\partial \tilde{U}^T(t)}{\partial U(t)} \frac{\partial \tilde{U}(t)}{\partial U(t)} \Big|_{U(t)=U^i(t)} \end{aligned} \quad (11)$$

where $B^{(i)}$ specifies the approximation of the inverse Hessian and $G[U^{(i)}(t)]$ is the gradient of the *J* with respect to the control inputs. The most popular formula known as Broyden-Fletcher-Goldfarb-Shanno (BFGS) algorithm to approximate the inverse Hessian is used here[8]. The proposed scheme of implementing the NNMPC is shown in Figure 2.

*Time Series Prediction with Neural Networks*

The purpose of our neural network model is to do time series prediction of the plant output. Given a series of control signals $\tilde{u}$ and past data $y_t$ it is desired to predict the plant output series $y_N$. The network is trained to do *one step ahead* prediction[9], i.e. to predict the plant output $y_{t+1}$ given the current control signal $u_t$ and plant output $y_t$. The neural network will implement the function

$$\hat{y}_{t+1} = f(u_t, y_t) \quad (12)$$

As it is discussed above, $y_t$ has to contain sufficient information for this prediction to be possible. It is assumed that $y_t$ is multivariable. One problem is that this method will cause a rapidly increasing divergence due to accumulation of errors. It therefore puts high demands on accuracy of the model. The better the model matches the actual plant the less significant the accumulated error. A sampling time as large as possible is an effective method to reduce the error accumulation as it effectively reduces the number of steps needed for a given time horizon. The neural network trained to do one step ahead prediction will model the plant. The acquisition of this model is also referred to as System Identification.

## IV. MODELING OF NEURAL NETWORK PREDICTIVE CONTROL (NNPC)

The three steps involved in the ANN model development are

### A. Generation of Input-Output data

The data generated to train the network should contain all the relevant information about the dynamics of the CSTR. The input was given to the conventional model of the CSTR and from the conventional model, the input and output were sampled for 0.02 sampling instants and the required sampled data are obtained to train the network.

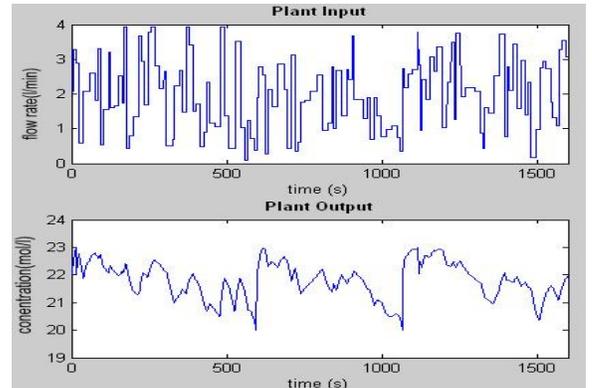

Figure 2: Input-Output Sequence

## B. Neural Network Architecture

The feed forward network with sigmoidal activation function was chosen based on the trials with different structures of multilayer perceptron.

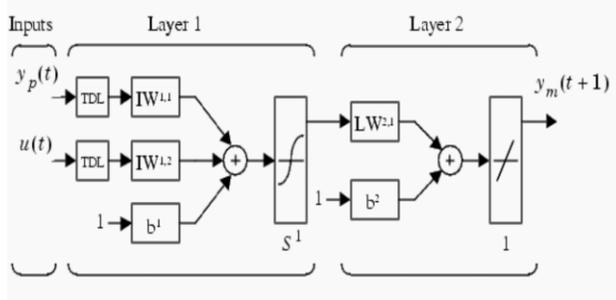

Figure 3: ANN model of the CSTR

The lowest error corresponds to 7 neurons in the hidden layer. Hence it is selected as optimal architecture of ANN. The ANN selected here consists of 4 neurons in the input layer, 7 neurons in the hidden layer and one neuron in the output layer. The ANN architecture used in the present work is shown in Figure 3. The training algorithm used in the CSTR modeling is back propagation algorithm. Before training the process weights are initialized to small random numbers. The weights are adjusted till error gets minimized for all training sets. When the error for the entire set is acceptably low, the training is stopped.

Table 2 shows the parameters used in developing the ANN model for the CSTR

| Parameters | Values |
|---|---|
| Input neurons | 4 |
| Output Neurons | 1 |
| Hidden layer Neurons | 7 |
| No. of hidden layer | 7 |
| Activation function | Sigmoidal |
| Training algorithm | Levenberg-Marquardt |
| Iteration | 10000 |
| Architecture | Feedforward |
| Initial weights | 1 |

*Table 2: ANN Parameters for CSTR modeling*

## C. Model Validation

The final step in developing the model is validation of the model [11]. Validation is performed by evaluating the model performance using trained data and test data. The input and target were presented to the network and the network was trained using Levenberg-Marquardt algorithm.

**Validation tests on training set:**

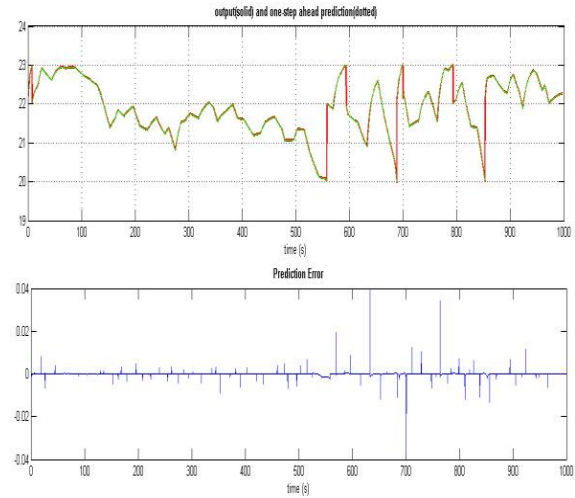

Figure 4: (a) One step ahead prediction of model, (b) Prediction error between model output and predicted output

**Validation tests on test set:**

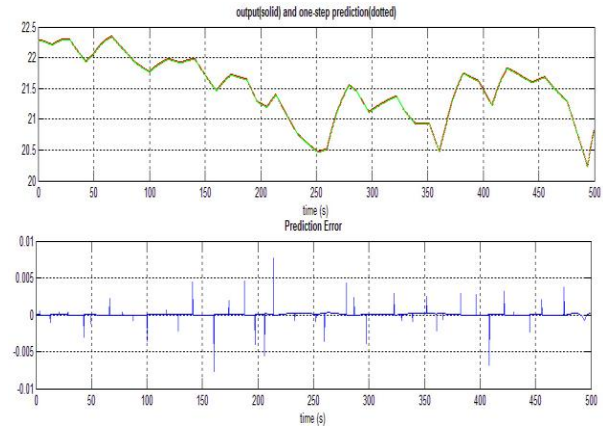

Figure 5 :(a) one step ahead prediction of model (validation set), (b) Prediction error between model output and predicted output (validation set)

## V. CONTINUOUS STIRRED TANK REACTOR

The Continuous Stirred Tank Reactor [6] is shown in Figure 6. This CSTR model in used as the nonlinear system.

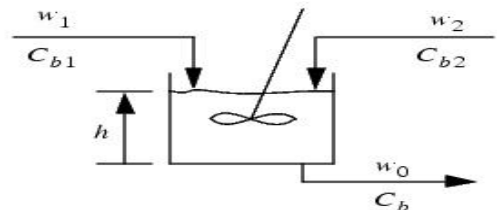

Figure 6: Continuous Stirred Tank Reactor

The equations which shows the dynamic model of the system is

$$\frac{dh(t)}{dt} = w_1(t) + w_2(t) - 0.2\sqrt{h(t)} \quad (14)$$

$$\frac{dC_b(t)}{dt} = (C_{b1} - C_b(t))\frac{w_1(t)}{h(t)} + (C_{b2} - C_b(t))\frac{w_2(t)}{h(t)} - \frac{k_1 C_b(t)}{(1+k_2 C_b(t))^2} \quad (15)$$

where h (t) is the liquid level, $C_b(t)$ is the product concentration at the output of the process, $w_1(t)$ is the flow rate of the concentrated feed $C_{b1}$ and $w_2(t)$ is the flow rate of the diluted feed $C_{b2}$. The input concentration are set to $C_{b1}$=24.9 and $C_{b2}$= 0.1.The constants associated with the rate of consumption are $k_1=k_2=1$.

The objective of the controller is to maintain the product concentration by adjusting the flow $w_1$ (t), $w_2$ (t) =0.1.The level of the tank h is not controlled. The designed controller uses a neural network model to predict future CSTR responses to potential control signals. The training data were obtained from the nonlinear model of CSTR.

## VI. SIMULATION RESULTS AND CONCLUSION

The objective of the control strategy is to govern the CSTR dynamics to force the system concentration to track a certain set-points. In this system, the input is the coolant flow rate and the output is the concentration of the process [12]. The identifier is trained and initialized before the control action starts. The input vector of the identifier includes coolant flow rates at different time steps (the sampling time is 20sec).
The performance of the proposed controller is shown in Figure 7. Evidently, the concentration values of the plant could track the set-point values excellent. It is to be noted that to improve the transient response, one may consider a larger prediction time. It is remarkable to note that because of highly nonlinearity nature of CSTR process, using the conventional control technique could not reach the control task. It can be seen in figure 7 that controller output is tracking the reference signal.

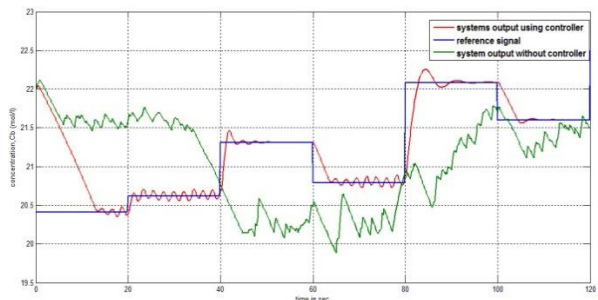
Figure 7: Response graph with and without controller

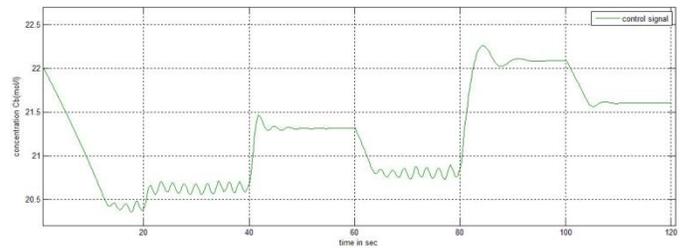
Figure 8: control signal by the controller

In this paper modeling of CSTR has been implemented using artificial neural networks. The neural model has been trained using data set obtained from dynamic equations. Feed forward neural network has been used to model the CSTR. The neural model has been designed as a black box model. The simulation results from conventional model and the neural model were compared for the given input variations and the results have been found satisfactory. The simulation shows that implementation of the Neural Network based advanced controllers for the set-point tracking case were able to force process output variables to their target values smoothly and within reasonable rise and settling times.

stability design", Automatica, vol 32, no. 12, 1701-1706, 1996.
[10]     Dan, W.P., 1996. Artificial Neural Networks- Theory and Applications. Prentice Hall, Upper Saddle River, New Jersey, USA.
 [11]     S.A.Billings, and W.S.F. Voon, "Correlation based model validity tests for nonlinear models. International Journal of Control, 44, 235–244.1986.

AUTHORS PROFILE

Author is presently working as Assistant Professor in Electrical and Electronics Department of Takshshila Institute of Engineering and Technology. He received the Masters degree in Electrical Engineering with specialization in Control Systems Engineering from Jabalpur Engineering College. His area of specialization is in Neural Networks, Control Systems, Fuzzy Logic and Artificial Intelligence.